\documentclass[conference, 10pt]{IEEEtran}
\usepackage[letterpaper, left=1.05in, right=1.05in, bottom=1.05in, top=0.75in]{geometry}
\IEEEoverridecommandlockouts

\usepackage{cite}
\usepackage{amsmath,amssymb,amsfonts}
\usepackage{algorithmic}
\usepackage{hyperref}
\usepackage{graphicx}
\usepackage{textcomp}
\usepackage{xcolor}
\usepackage[capitalise]{cleveref}
\usepackage{flushend}
\usepackage{booktabs}
\usepackage[toc]{glossaries}
\newacronym{cnn}{CNN}{Convolutional Neural Network}
\newacronym{wdn}{WDN}{Water Distribution Network}
\newacronym{dma}{DMA}{District Metered Area}
\newacronym{dd}{DD}{Detection Delay}
\newacronym{mse}{MSE}{Mean Squared Error}
\newacronym{tp}{TP}{True Positive}
\newacronym{tn}{TN}{True Negative}
\newacronym{fp}{FP}{False Positive}
\newacronym{fn}{FN}{False Negative}
\newacronym{mnf}{MNF}{Minimum Night Flow}
\newacronym{relu}{ReLU}{Rectified Linear Unit}
\newacronym{nrw}{NRW}{Non Revenue Water}
\newacronym{ml}{ML}{Machine Learning}
\newacronym{iot}{IoT}{Internet of Things}
\newacronym{hdpe}{HDPE}{High Density Polyethylene}
\newacronym{frf}{FRF}{Frequency Response Function}
\newacronym{psd}{PSD}{Power Spectral Density}
\newacronym{sf}{SF}{Spectral Flatness}
\newacronym{sro}{SRO}{Spectral Roll-Off}
\newacronym{sb}{SB}{Spectral Bandwidth}
\newacronym{zcr}{ZCR}{Zero-Crossing Rate}
\newacronym{sc}{SC}{Spectral Centroid}
\newacronym{rms}{RMS}{Root Mean Square}
\newacronym{ol}{OL}{Orifice Leak}
\newacronym{lc}{LC}{Longitudinal Crack}
\newacronym{cc}{CC}{Circumferential Crack}
\newacronym{gl}{GL}{Gasket Leak}
\newacronym{nl}{NL}{No-Leak}
\newacronym{pvc}{PVC}{Polyvinyl Chloride}
\newacronym{mems}{MEMS}{Micro-Electro-Mechanical-System}
\newacronym{sfs}{SFS}{Sequential Feature Selector}
\newacronym{lr}{LR}{Linear Regression}
\newacronym{logr}{LogR}{Logistic Regression}
\newacronym{rf}{RF}{Random Forest}
\newacronym{gb}{GB}{Gradient Boosting}
\newacronym{svm}{SVM}{Support Vector Machines}
\newacronym{mfcc}{MFCC}{Mel-Frequency Cepstral Coefficients}
\newacronym{lstm}{LSTM}{Long Short-Term Memory}
\newacronym{rnn}{RNN}{Recurrent Neural Network}
\newacronym{bliff}{BLIFF}{Burst Location Identification Framework}
\newacronym{snr}{SNR}{Signal to Noise Ratio}
\newacronym{fpr}{FPR}{False Positive Rate}
\newacronym{hdbscan}{HDBSCAN}{Hierarchical Density-Based Spatial Clustering of Applications with Noise}

\def\f{\mathbf{f}}

\def\p{{{p}}}
\def\P{{\mathbf{P}}}

\def\W{{\mathbf{W}}}
\def\hatW{{\hat{\mathbf{W}}}}
\renewcommand{\subsection}[1]{\vspace{.5em}\noindent\textbf{#1}.}

\def\BibTeX{{\rm B\kern-.05em{\sc i\kern-.025em b}\kern-.08em
    T\kern-.1667em\lower.7ex\hbox{E}\kern-.125emX}}

\begin{document}

\title{Enhanced Water Leak Detection with Convolutional Neural Networks and One-Class Support Vector Machine
{
}}

\iftrue
    \author{\IEEEauthorblockN{Daniele Ugo Leonzio}
    \IEEEauthorblockA{
    \textit{Politecnico di Milano}\\
    Milan, Italy \\
    danieleugo.leonzio@polimi.it}
    \and
    \IEEEauthorblockN{Paolo Bestagini}
    \IEEEauthorblockA{
    \textit{Politecnico di Milano}\\
    Milan, Italy \\
    paolo.bestagini@polimi.it}
    \and
    \IEEEauthorblockN{Marco Marcon}
    \IEEEauthorblockA{
    \textit{Politecnico di Milano}\\
    Milan, Italy \\
    marco.marcon@polimi.it}
    \and
    \IEEEauthorblockN{Stefano Tubaro}
    \IEEEauthorblockA{
    \textit{Politecnico di Milano}\\
    Milan, Italy \\
    stefano.tubaro@polimi.it}
    }
\else
\author{\IEEEauthorblockN{Anonymous Authors} }
\fi

\maketitle

\begin{abstract}
Water is a critical resource that must be managed efficiently. However, a substantial amount of water is lost each year due to leaks in \glspl{wdn}. This underscores the need for reliable and effective leak detection and localization systems. In recent years, various solutions have been proposed, with data-driven approaches gaining increasing attention due to their superior performance.  In this paper, we propose a new method for leak detection. The method is based on water pressure measurements acquired at a series of nodes of a \gls{wdn}. Our technique is a fully data-driven solution that makes only use of the knowledge of the \gls{wdn} topology, and a series of pressure data acquisitions obtained in absence of leaks. The proposed solution is based on an feature extractor and a one-class \gls{svm} trained on no-leak data, so that leaks are detected as anomalies. The results achieved on a simulate dataset using the Modena \gls{wdn} demonstrate that the proposed solution outperforms recent methods for leak detection.
\end{abstract}


\section{Introduction}
\label{sec:intro}

Between the growing threats of climate change and pollution, the shift to an environmentally sustainable economy is gaining momentum. This transition highlights the need to understand the environmental impact of our actions. Key strategies include transforming energy production, reducing food waste, and achieving carbon negativity, all aimed at cost-effective sustainability with a positive environmental footprint.

A notable trend is the increased monitoring of critical infrastructure to ensure better working conditions, reduce waste, and boost ecological benefits. Monitoring helps anticipate issues, implement preventive maintenance, and optimize production, enhancing operational efficiency and environmental contributions.

In \glspl{wdn}, a major concern is water leakage. Factors such as rising demand, aging infrastructure, and environmental degradation have worsened water scarcity in recent years. Leaks, which account for about 30\% of urban water usage, have significant economic and environmental impacts \cite{Mounce2010}. Rapid and reliable leak detection in pipe networks offers considerable advantages, reducing operational costs, improving service levels for water utilities, and preventing water pollution. Detecting leaks is not just an economic issue but also an environmental and safety imperative.

In response to these challenges, modern electronic and information technologies have enabled the development of advanced sensing systems for water leak detection \cite{Mashford2012, Leonzio2024Smartwater, ADACHI2014, Mashhadi2021, IROFTI2020Fault, Leonzio2024ICASSP}. 

To address the limitations of hardware-based approaches, data-driven methods have gained prominence in recent years. These approaches leverage statistical analysis and machine learning algorithms to identify and locate leaks within \glspl{wdn} without the need for detailed pipe-specific information \cite{Chan2018}. Data-driven solutions prioritize the utilization of broad network topology information to pinpoint the location of leaks. They effectively bypass the requirement for detailed pipe parameters such as diameter, material composition, and thickness. For instance, data-driven methods such as \gls{svm} have been employed to solve the leak detection problem based on pressure sensor measurements and transient data \cite{Mashford2012, Perez2014}.  Essentially, these approaches learn from historical data using statistical or pattern recognition tools \cite{Romano2014AutomatedDO}, endowing them with a robust generalization capability. Generalization allows them to be effectively applied across a diverse range of network configurations, regardless of the specific details of individual pipes. In simpler terms, a data-driven solution developed for one particular \gls{wdn} can, with its inherent adaptability, be successfully implemented in other networks with entirely different pipe characteristics.

In recent years, deep learning approaches have gained traction due to their effectiveness in leak detection. These techniques have revolutionized the field by providing advanced frameworks for handling complex patterns and variabilities. Quiñones-Grueiro et al. \cite{Quinones-Grueiro2021} integrated deep neural networks with Gaussian process regression to detect and localize leaks, demonstrating the potential of combining data-driven and model-based approaches. The \gls{bliff} framework utilizes pressure data in a mixed data-driven and model-based approach, a method less explored in other studies to pinpoint leaks more accurately by focusing on specific pipes rather than general areas. To address the complexity of pressure signals, linear connections were used in place of the typical convolutional layers in DenseNet \cite{Zhang2019}, simplifying calculations. The work \cite{Geelen2019} by Geelen et al. explores the use of pressure sensor data to enhance the monitoring and management of water distribution systems. The authors develop a framework, called Monitoring Support, that utilizes real-time pressure data to detect anomalies and improve the system's overall reliability. The detection is done using a moving window range statistic and the  \gls{hdbscan} method \cite{mcinnes2017hdbscan}.

Soldevilla et al. \cite{Soldevila_2022} propose a method that detects leaks by a sequential monitoring algorithm that analyzes the inlet flow of a \gls{dma}. A \gls{dma} is a portion of a more complex \gls{wdn}. This sectorization is done through valves or disconnection of network pipes with inlet and outlet flow metered \cite{Galdiero2016DecisionSS}. They formulated the leak detection problem as a change-point detection problem, solved by an ad hoc two layer algorithm including a hypothesis test to validate each detection and estimate the leak size and leak time.

A novel deep learning approach is proposed in \cite{Leonzio2023}. In this work the authors propose to solve the leak detection and localization problem using an autoencoder trained on leak free samples. While the authors achieved improvements in \gls{dd} and localization error, their work is limited to a small network that is not representative of a real-world \gls{wdn}. 

In this work we propose a new method to detect leak in a \gls{wdn}. The proposed method analyzes measurements of water pressure at a series of nodes of a \gls{wdn}. In particular, we train a feature extractor on top of measurements obtained in case of no leaks, as done in \cite{Leonzio2023, Leonzio2023IECON}. After with the feature extracted from the leak free samples we train a one-class \gls{svm}. At test time, the one-class \gls{svm} detects leaks as anomalies. The results achieved are promising, as we obtain an average \gls{dd} of about 40.21 hours (i.e., we detect a leak after 40.21 hours). This improvement was demonstrated on the Modena \gls{wdn}, a real-size network, underscoring the practical applicability of the proposed approach.

The paper is organized as follows. \cref{sec:prob_and_method} reports the formal description of the problem addressed in this work and describes the main steps of the proposed technique. \cref{sec:ex_setup} focuses on presenting the dataset, the autoencoder training, and how we select the parameters of our method. \cref{sec:results} presents
the performance results on leak detection.
Finally, \cref{sec:conclusion} concludes the paper.



\section{Water Leak Detection}
\label{sec:prob_and_method}

\subsection{Problem Formulation}
In our work we propose a method for water leak detection in a \gls{wdn} by monitoring the pressures values measured at a series of nodes. 

Let us consider a \gls{wdn} composed by $K$ nodes. Each node is equipped with a pressure sensor.
The pressure measured at the $k$-th node is a time series sampled at regular intervals. The $n$-th sample at the $k$-th node is defined as $\p_k(n)$. 

The presence of a leak modifies the pressure $\p_k$ with a drop of the pressure value at the sample $N^\star$ corresponding to the leak starting time $T^\star$, with a reduction that is proportional to the leak size. 

With this setup in mind, we can define the two main goals of this paper as follows
\begin{itemize}
    \item \textit{To detect if a leak starts in a \gls{wdn}}: this means attributing to the \gls{wdn} under analysis a label $\hat{c}$ equals to 1 if a leak is detected or 0 otherwise.
    \item \textit{To determine the leak starting time}: this means to compute $\hat{T}$ which is an estimate of $T^\star$.
    
\end{itemize}

\subsection{Proposed Method}
We propose solving the leak detection problem using an algorithm based on a feature extractor and a one-class \gls{svm}. The feature extractor is derived from the encoder section of an autoencoder, a specialized type of neural network designed to encode input into a compressed and meaningful representation, then decode it back so that the reconstructed input closely resembles the original \cite{Bank2020}. The autoencoder is trained on data representing a no-leak scenario, allowing the system to encode the input pressure matrix into a compact vector. By training with no-leak samples, we ensure that anomalies are not introduced during the training phase. The idea of using an autoencoder for leak detection in a \gls{wdn} trained on no-leak samples has already been explored in the literature \cite{Leonzio2023}. In this work, we use the autoencoder proposed in \cite{Leonzio2023} as our starting point. The one-class \gls{svm} is trained on the embeddings extracted from the no-leak samples, in order to spot the embeddings from leak scenarios as anomalies. The feature extraction step is essential because training a one-class \gls{svm} directly on the raw input data can be less effective due to the high dimensionality and complexity of the data. The encoder compresses and refines the input into a more meaningful and compact representation, making it easier for the one-class \gls{svm} to accurately detect anomalies. 

\begin{figure}[t]
    \centering
    \includegraphics[width= 0.95\columnwidth]{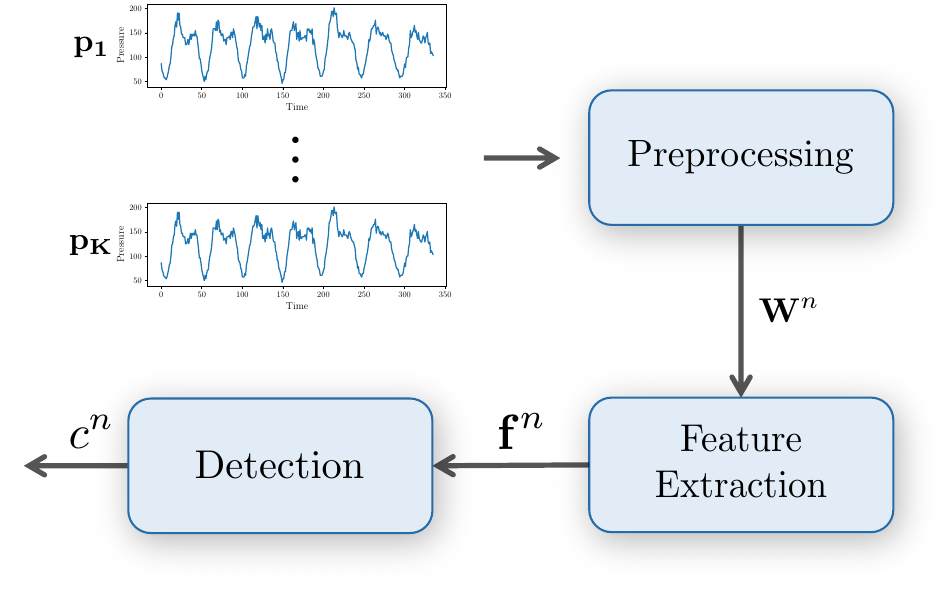}
    \caption{Block diagram of the proposed method.}
    \label{fig:pipeline}
    \vspace{-1em}
\end{figure}

As shown in \cref{fig:pipeline}, our pipeline can be broken down into 3 main blocks: \textit{Preprocessing}; \textit{Feature Extraction}; \textit{Detection}.

In the following, we report all the details related to each one of these blocks.

\subsubsection{Preprocessing}
Preprocessing is a crucial step for preparing the data before it is input into the feature extractor. During this stage, the system receives a new sample $\p_k(n)$ from each node simultaneously at every time instant. The preprocessing component collects the most recent $L$ samples from all $K$ sensors and arranges them into a fixed-size matrix $K \times L$, which will then be processed by the feature extractor in the next phase.

This component operates as a circular buffer, continuously updating by discarding the oldest sample whenever a new one is obtained. This method ensures that the analysis is consistently performed on a fixed-length window, concentrating solely on the most recent data.

At each sampling instant $n$, the preprocessing step selects a window of $L$ samples from $\p_k$, comprising the most recent $L-1$ samples along with the latest one.

Since data from all nodes are processed simultaneously, the matrix $\W^n$
is formed by concatenating the last $L$ pressure samples collected from all nodes within the \gls{wdn}, as shown:

\begin{equation}
\W^n = 
\left[
\begin{array}{@{\hskip 4pt}c@{\hskip 4pt}c@{\hskip 4pt}c@{\hskip 4pt}c@{\hskip 4pt}}
     \p_1(n\!-\!L\!+\!1) & \ldots & \p_1(n\!-\!1) & \p_1(n) \\
     \p_2(n\!-\!L\!+\!1) & \ldots & \p_2(n\!-\!1) & \p_2(n) \\
     \vdots              & \vdots & \vdots       & \vdots \\
     \p_K(n\!-\!L\!+\!1) & \ldots & \p_K(n\!-\!1) & \p_K(n) \\
\end{array}
\right], 
\end{equation}
where the $k$-th row represents measurements from the $k$-th node, and $K$ is the number of nodes in the \gls{wdn}.

\subsubsection{Feature Extraction}
\label{sec:feat_extractor}
The second step of our method consists in analyzing the input matrix $\W^n$ with a feature extractor in order to have a compact and meaningful representation of it.
The feature extractor is obtained as the encoder part of an autoencoder.
The autoencoder is trained to reconstruct the input signal in case of absence of leak. In this way, the training process of the autoencoder is not affected by possible anomalies in the input matrix  $\W^n$. The complete description of the autoencoder and its training process can be found in \cref{sec:autoencoder}.

The feature extractor takes the input $\W^n$ with size $K \times L$, and computes a reduced dimensionality version of it.
This is composed by 2 convolutional layers:
\begin{itemize}
    \item The first layer is a 1D Convolutional Layer with 64 filters of size 7 and stride 2, followed by a \gls{relu}.
    \item The second layer is a 1D Convolutional Layer with 32 filters of size 7 and stride 2, followed by a \gls{relu}.
\end{itemize}

The output of the feature extractor model is vector $\f^n$, which represent the reduce dimensionality version of the input matrix $\W^n$. 

\begin{equation}
    \f^n = \mathcal{E}(\W^n).
\end{equation}

\subsubsection{Detection}
\label{sec:detection}
The detection step of the proposed method is based on a one-class \gls{svm}.
The one-class \gls{svm} approach is a machine learning method used for anomaly detection in scenarios where the training data consists solely of instances from one class, typically representing normal behavior. This model learns the boundary that best encompasses the normal data, mapping it into a high-dimensional feature space where it defines a hyperplane or a boundary around the majority of the data points. During testing, the one-class \gls{svm} classifies new instances based on their proximity to this boundary: data points that lie within the boundary are classified as normal, while those that fall outside are flagged as anomalies or outliers \cite{scholkopf2001estimating}. This approach is especially useful in scenarios where anomalous data is rare or challenging to collect for training, such as in the case of water leak detection. 

The one-class \gls{svm} has been trained using the no-leak scenario embeddings from the same samples used to train the autoencoder. 

The output of the one-class \gls{svm} is a soft score $\hat{y}^n$. 

To analyze the behavior over time, we concatenate all the predicted values $\hat{y}$ for each analyzed window, forming a score vector denoted as $\mathbf{\hat{y}}$. To suppress noise and patterns unrelated to anomalies, we apply a moving average filter to the vector $\mathbf{\hat{y}}$ obtaining the vector $\mathbf{\hat{y}}_{\text{smoothed}}$. Finally, we apply a threshold $\gamma$ to the smoothed vector, designating all values exceeding this threshold as anomalies.

Formally,
\begin{equation}
    \hat{c}^n = 
    \begin{cases}
        1, & \text{if} \;  \hat{y}^n_{\text{smoothed}}  \geq \gamma,\\
        0 , & \text{otherwise}.
    \end{cases}
    \label{eq:c}
\end{equation}

Here, $\hat{c}^n$ represents the binary anomaly indicator at the sample $n$, where values in the smoothed vector $\mathbf{\hat{y}}_{\text{smoothed}}$  above the threshold $\gamma$ are set to 1.

If a leak is detected (i.e., $\hat{c}^n=1$ for at least one value of $n$), the time instant associated to the last received pressure sample is set as leak starting time instant. We denote the position of this sample as $\hat{N}$, which corresponds to the time instant $\hat{T}$.

\section{Experimental Setup}
\label{sec:ex_setup}

\subsection{Dataset}
To develop and test our algorithm, we utilized the Modena \gls{wdn}, a network frequently employed as a benchmark in various studies \cite{Suribabu_2019_analysis, monsef2019comparison, quinones2021robust}. The network comprises 268 nodes and 317 pipes, with no pumps or valves included in the model. Each pipeline in the network is characterized by its diameter, length, and roughness. The pipe lengths range from 1 m to 1094.73 m, while diameters vary between 100 mm and 400 mm.

We simulated 500 distinct scenarios, including both leak and no-leak cases. In the leak scenarios, we varied the leak positions and sizes, while all scenarios shared the same fixed network topology. The simulations were conducted using EPANET \cite{Rossman2000}. To further increase the dataset's complexity, we introduced randomness to the diameter, length, and roughness values of each pipe.

\begin{figure}[t]
    \centering
    \includegraphics[width = 0.65 \columnwidth]{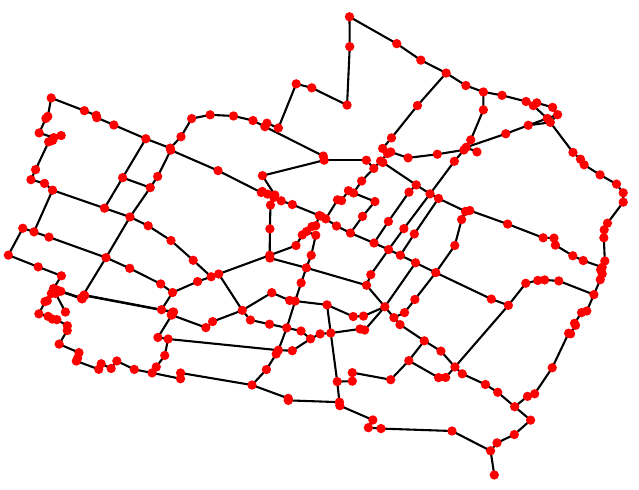}
    \caption{Modena network topology. Each dot represents a node. Lines represent connections between nodes.}
    \label{fig:network_example}
     \vspace{-1em}
\end{figure}

In \cref{fig:network_example} we report the topology of the Modena network used in our work.

In each scenario, we have a simulation of a \gls{wdn} sampled every 30 minutes for one year, so every 30 minutes a new sample is added and the buffer is shifted. The node number $K$ is equal to 268 for all scenarios. For every node we have the pressure time series.

\subsection{Autoencoder}
\label{sec:autoencoder}
The autoencoder can be divided in two parts: the \textit{Encoder} ($ \mathcal{E}$) and the \textit{Decoder} ($\mathcal{D}$).

The encoder takes the input $\W^n$ with size $K \times L$, and computes a reduced dimensionality version of it. The encoder layers are described in \cref{sec:feat_extractor}
The decoder takes the low-dimensionality output of the encoder as input, and estimates the output $\hat{\W}^n$ with size $K \times L$.
This is composed by 3 convolutional layers:
\begin{itemize}
    \item The first layer is a 1D Transposed Convolutional Layer with 32 filters of size 7 and stride 2, followed by a \gls{relu}.
    \item The second layer is a 1D Transposed Convolutional Layer with 64 filters of size 7 and stride 2, followed by a \gls{relu}.
    \item The third layer is a 1D Transposed Convolutional Layer with 32 filters of size 7 and stride 1.
\end{itemize}

Both the encoder and the decoder parameters have been selected after a grid search procedure, and the structure proposed here is the best in terms of loss value and computational cost.

We train the autoencoder on no-leak data considering \gls{mse} as loss function.
The output of the autoencoder is $\hatW^n$, a reconstruction of the input matrix $\W^n$ obtained as
\begin{equation}
    \hatW^n = \mathcal{AE}(\W^n),
    \label{eq:autoencoder}
\end{equation}
where $\mathcal{AE}$ implements the autoencoder operator.
The \cref{eq:autoencoder} can be written also as 
\begin{equation}
    \hatW^n = \mathcal{D}(\mathcal{E}(\W^n)),
    \label{eq:autoencoder_split}
\end{equation}
where the autoencoder model $\mathcal{AE}$ has been split in to its encoder $\mathcal{E}$ and decoder part $\mathcal{D}$.

\subsection{Parameters selection}
In this section we report additional details about the parameters we adopt in our pipeline. 

The first parameter is the sample length $L$ of the window. We set $L$ in order to cover one week of measurements within each sliding window. This means that $L=336$ samples considering the simulation sampling rate. Thanks to this choice we are able to model both daily and weekly periodicity, while still remaining robust to seasonal variations because these are slower than the window size and window update. This parameter set automatically also the setup time of the algorithm, which is equal to the window length (i.e., one week). The setup time refers to the initial period required for the algorithm to start functioning effectively, during which it accumulates enough data to populate the first window.

The second parameter is the threshold $\gamma$ reported in \cref{sec:detection}. The $\gamma$ value has been chosen comparing the results obtained with different values of \gls{fpr}, on a validation set formed by 50 simulations. In our method we decide to adopt a \gls{fpr} of 10\% which gave as a threshold value of 7.44.
\section{Results}
\label{sec:results}

This section presents the results we achieved with our method on Modena network dataset. We compare our detection results against \cite{Leonzio2023} and a refined version of \cite{Leonzio2023}, in which we change the original threshold proposed by the authors in order to adapt the method on the different data used in this work. In addition we test the performance of the proposed method in case of noisy data with different \gls{snr} levels and the generalization capabilities on a different \gls{wdn}. 

\subsection{Detection}
The performance of the leak detection system was evaluated using accuracy as the primary metric. A true positive is defined as a case where a leak is both present and detected, while a true negative corresponds to a scenario where a leak is not present and not detected. The accuracy has been computed considering the $\hat{c}$ results with respect the true label $c$. This metric is a common state-of-the-art metric for measuring the leak detection performances \cite{Soldevila_2022}.

The results obtained by the baseline and the proposed methods are presented in \cref{tab:det_accuracy}. Notably, without tuning the threshold for the new dataset, the baseline method \cite{Leonzio2023} yielded a relatively low accuracy. However, by adjusting the threshold, the number of detected leaks significantly increased, reaffirming \cite{Leonzio2023} as a viable solution for addressing the leak detection problem.


\begin{table}[tb]
\centering
\caption{Detection Accuracy}
\begin{tabular}{@{}c|c@{}}
\toprule
Method & \textbf{Accuracy} \\
\midrule

Baseline \cite{Leonzio2023} & 0.22 \\
Baseline \cite{Leonzio2023} refined  & 0.80 \\
Proposed Method  & 0.92\\
\bottomrule 
\end{tabular}
\label{tab:det_accuracy}
\vspace{-1em}
\end{table}

\subsection{Leak Starting Time}
To measure the performance for the starting time detection we used as metric the \glsfirst{dd}. 
The \gls{dd} is defined as
\begin{equation}
  \label{eq:detdelay}
  \mathrm{DD} = \hat{T} - T^\star,
\end{equation}
were we assume that $\hat{T} \geq T^\star$ as we consider the detected point before the leak happens as false positive. The \gls{dd} is expressed in hours. With our pipeline we were able to achieve a \gls{dd} average value of 40.21 hours. 

The comparison with the baseline method \cite{Leonzio2023} is provided in \cref{tab:det_delay}. The \gls{dd} achieved by the refined baseline is comparable to other state-of-the-art methods, despite being higher than the value reported in \cite{Leonzio2023}. In contrast, our proposed method achieves a detection delay that is an order of magnitude smaller than the baselines.

\begin{table}[tb]
\centering
\caption{Detection Delay in hours}
\begin{tabular}{@{}c|c@{}}
\toprule
Method & \textbf{Detection Delay} \\
\midrule

Baseline \cite{Leonzio2023} & 71.00 \\
Baseline \cite{Leonzio2023} refined  & 61.64 \\
Proposed Method  & 40.21\\
\bottomrule 
\end{tabular}
\label{tab:det_delay}
\vspace{-1em}
\end{table}

\subsection{Noise Robustness}
To check the robustness to possible noisy data input to the algorithm, we simulated Gaussian noise at different \gls{snr} levels, which we summed to the various time series before the \textit{Preprocessing} step. We want to highlight that the proposed method blocks were trained using only noise free signal and the noisy data were used only in test phase. 

The results achieved are shown in \cref{tab:noise_res}. It is possible to see that for high value of \gls{snr} we are still able to achieve good performances both in accuracy and \gls{dd}. By reducing the \gls{snr}, a marked decline in performance is observed. To address this issue, potential strategies include incorporating a noise reduction stage at the initial phase of the proposed method’s pipeline or training the feature extractor and one-class \gls{snr} with noisy data to enhance robustness.

\begin{table}[tb]
\centering
\caption{Detection Accuracy and Delay for different \gls{snr} levels}
\begin{tabular}{@{}c|c|c @{}}
\toprule
\gls{snr} & \textbf{Accuracy} & \textbf{Detection Delay}\\
\midrule

45 dB & 0.72 & 81.4 \\
40 dB & 0.63 & 187.1\\
35 dB & 0.63 & 321.5\\
30 dB & 0.59 & 422.2\\
25 dB & 0.59 & 454.6\\
\bottomrule 
\end{tabular}
\label{tab:noise_res}
\vspace{-1em}
\end{table}

\subsection{Detection on Different \glspl{wdn}}
To evaluate the generalization capabilities of the proposed method on previously unseen \glspl{wdn}, we applied it to the Hanoi and Pescara networks. These experiments utilized the LeakDB dataset \cite{vrachimis2018leakdb}, which provides 500 simulated scenarios, including both leak and no-leak cases, for the Hanoi network. Since the Hanoi \gls{wdn} has fewer nodes than the network used during the method's development, we replicated the nodes to match the input size required by the feature extractor. The accuracy and \gls{dd} values obtained are reported in \cref{tab:hanoi_res}.  

The results are promising, particularly as the method was applied without significant adaptation to the specifics of the new network, except for adjusting the number of nodes. The method successfully detected 77\% of leaks in the Hanoi network. While the \gls{dd} value (102 hours) is slightly higher than those reported for state-of-the-art methods, this could be improved by optimizing the detection threshold for this specific network.  

\begin{table}[tb]
\centering
\caption{Detection Accuracy and Delay for Hanoi \gls{wdn}}
\begin{tabular}{@{}c|c|c @{}}
\toprule
\gls{wdn} & \textbf{Accuracy} & \textbf{Detection Delay (hours)} \\
\midrule
Hanoi & 0.77 & 102 \\
\bottomrule 
\end{tabular}
\label{tab:hanoi_res}
\vspace{-1em}
\end{table}

Similarly, we applied the proposed method to the Pescara \gls{wdn}. The Pescara network consists of 68 junctions, 3 reservoirs, and 99 pipes. We simulated 500 scenarios, both with and without leaks, varying the leak positions and sizes for the leak scenarios. As with the Modena network, the simulations were conducted using the EPANET software, with data sampled every 30 minutes.  

The results for the Pescara \gls{wdn} are shown in \cref{tab:pescara_res}. The proposed method achieved an accuracy of 82\% and a mean \gls{dd} of 54 hours. These results are closer to those obtained for the Modena network, further highlighting the generalization capabilities of the proposed pipeline across different network topologies.  

\begin{table}[h]
    \centering
    \caption{Detection Accuracy and Delay for Pescara \gls{wdn}}
    \begin{tabular}{@{}c|c|c@{}}
        \toprule
        \textbf{WDN} & \textbf{Accuracy} & \textbf{Detection Delay (hours)} \\
        \midrule
        Pescara & 0.82 & 54 \\
        \bottomrule
    \end{tabular}
    \label{tab:pescara_res}
    \vspace{-1em}
\end{table}

\section{Conclusion}
\label{sec:conclusion}

In this work, we propose a novel methodology for leak analysis in a \gls{wdn}. Specifically, we address two key tasks:
\begin{itemize}
\item Detecting the presence of a leak in a \gls{wdn}.
\item Determining the starting time of the leak.
\end{itemize}

The proposed methodology combines a feature extractor with a one-class \gls{svm} trained exclusively on no-leak data to identify leaks as anomalies.

To evaluate the method, we simulated various scenarios using the Modena \gls{wdn} as the backbone network. Our approach demonstrates strong performance in detecting leaks. The pipeline is highly efficient, achieving an average detection delay (\gls{dd}) of 40.21 hours with a detection accuracy of 92\%.

To our knowledge, this is one of the first studies to comprehensively address the challenges of noisy input data and generalization capabilities in the field of water leak detection. We extensively evaluated the robustness of the proposed method across different signal-to-noise ratio (\gls{snr}) levels and assessed its generalization on two additional \glspl{wdn} that were not part of the training phase. These results underscore the method's reliability and adaptability across varying network conditions.

For future work, we aim to validate the algorithm on real-world scenarios using actual pressure measurements. We also plan to integrate a leak localization component and develop advanced strategies to handle noisy input data more effectively, further enhancing the method's applicability in practical settings.

\bibliographystyle{IEEEtran}
\bibliography{biblio}

\end{document}